\documentclass[letterpaper, 10 pt, conference]{ieeeconf}  

\IEEEoverridecommandlockouts                              

\usepackage{latexsym}
\usepackage{amssymb}
\usepackage{amsmath}
\usepackage{booktabs}
\usepackage{graphicx}
\usepackage{color}
\usepackage{listings}
\usepackage{float}
\usepackage{url}
\makeatletter
\let\labelindent\@undefined
\makeatother
\usepackage{enumitem}

\title{\LARGE \bf Generating Actionable Robot Knowledge Bases\\ by Combining 3D Scene Graphs with Robot Ontologies}

\author{Giang Nguyen$^{1}$, Mihai Pomarlan$^{2}$, Sascha Jongebloed$^{1}$, Nils Leusmann$^{1}$, Minh Nhat Vu$^{3}$ and Michael Beetz$^{1}$
	\thanks{$^{1}$Institute for Artificial Intelligence,
		University of Bremen, Germany}%
	\thanks{$^{2}$Appliable Linguistics,
		University of Bremen, Germany}%
        \thanks{$^{3}$Automation \& Control Institute,
		TU Wien, Austria}%
}

\begin{document}

\newfloat{lstfloat}{htbp}{lop}
\floatname{lstfloat}{ing}
\def\lstfloatautorefname{Listing} 
\lstset{escapeinside={(*@}{@*)}}

\maketitle
\thispagestyle{empty}
\pagestyle{empty}

\begin{abstract}
In robotics, the effective integration of environmental data into actionable knowledge remains a significant challenge due to the variety and incompatibility of data formats commonly used in scene descriptions, such as MJCF, URDF, and SDF. This paper presents a novel approach that addresses these challenges by developing a unified scene graph model that standardizes these varied formats into the Universal Scene Description (USD) format. This standardization facilitates the integration of these scene graphs with robot ontologies through semantic reporting, enabling the translation of complex environmental data into actionable knowledge essential for cognitive robotic control. We evaluated our approach by converting procedural 3D environments into USD format, which is then annotated semantically and translated into a knowledge graph to effectively answer competency questions, demonstrating its utility for real-time robotic decision-making. Additionally, we developed a web-based visualization tool to support the semantic mapping process, providing users with an intuitive interface to manage the 3D environment.
\end{abstract}

\section{Introduction}

In AI-powered and cognition-enabled robotics, robot agents face the challenge of fulfilling underdetermined task requests such as "prepare a breakfast" or "bring me something to drink." To accomplish these tasks, robots must infer the specific body movements required, which heavily depend on the given environment and the robot's knowledge and reasoning capabilities. 
This knowledge includes the physics, geometry, and visual characteristics of the environment and its objects. Although the necessary details for computing these movements are contained within virtual reality environments' scene graph data structures, these structures are not standardised, inherently machine-understandable, or interpretable. This limitation restricts a robot's ability to answer task-critical queries in changing environments, such as whether milk is stored within a container, how to operate a refrigerator or the outcomes of handling a milk carton by the lid.

During the processing of scene graph data, a significant challenge arises from the diversity and vast volume of environmental data, which is distributed across various scene description formats in robotics such as URDF, MJCF, and SDF, each serving different purposes \cite{ivanou2021robot}. Such diversity complicates the development of generalized robotic control programs, as the necessary semantic information for interpreting the environment is often missing, requiring intensive manual semantic labeling. Additionally, the dynamic nature of environments demands that semantic interpretations be continuously updated; for example, food must be reclassified as waste once it is expired or inedible. 
Therefore, there is a pressing need for a pipeline to manage effectively 
and simplify 
this complexity.

\begin{figure}[t]
    \centering
    \includegraphics[width=\columnwidth, height=5.5cm]{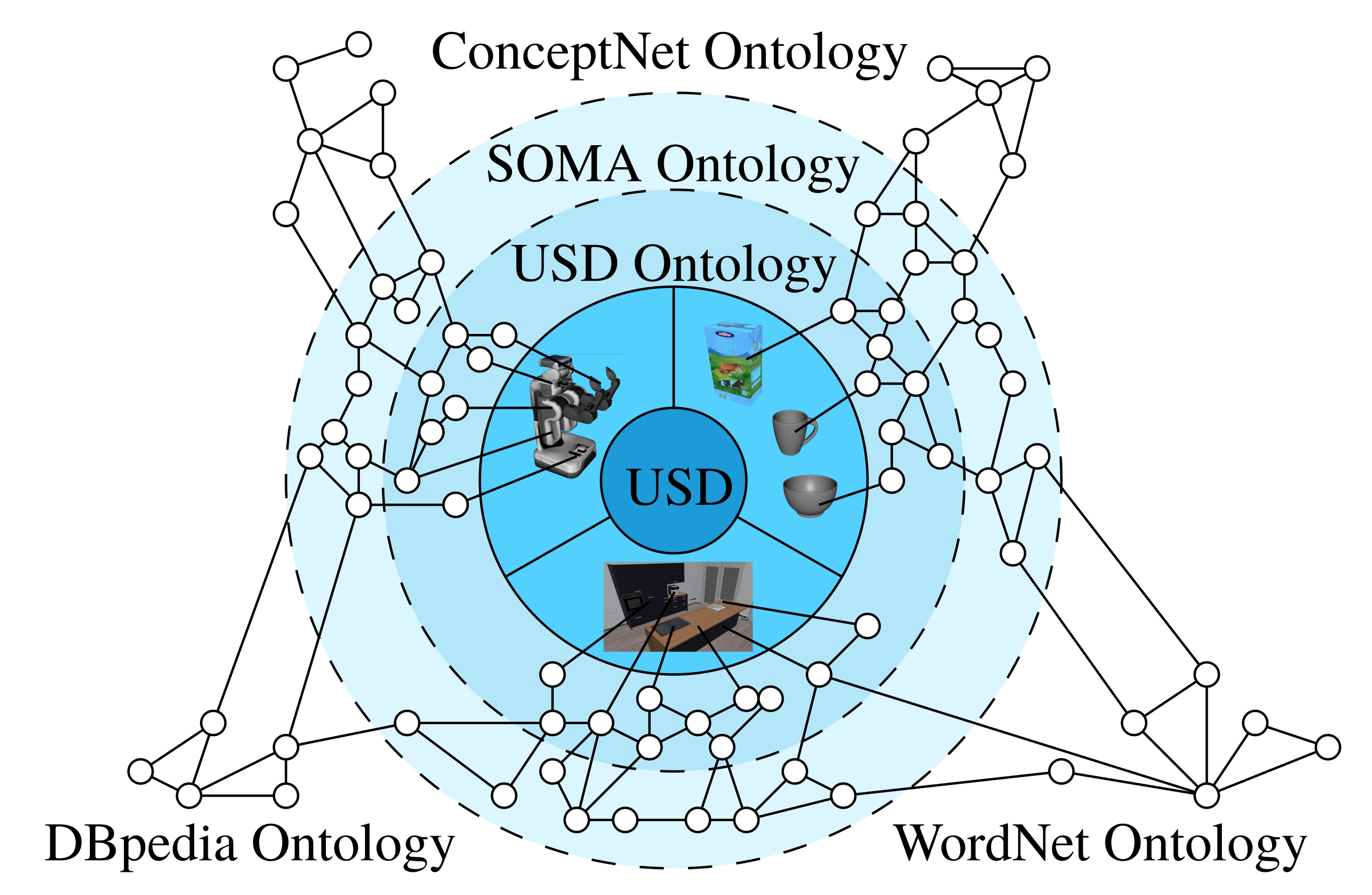}
    \caption{The generation of a Knowledge Graph from various ontologies derived from a USD Scene Graph.}
    \label{fig:VisualAbstract}
\end{figure}

In this paper, we explore the research hypothesis that automated linking of the nodes in scene graphs to a general-purpose robot ontology, actionable within robot agents, could provide the necessary knowledge to fulfill such vague task requests effectively, as shown in figure \ref{fig:VisualAbstract}.
To test this hypothesis, we import arbitrary 3D environments from ProcTHOR \cite{deitke2022procthor}, a framework capable of generating an unlimited variety of virtual environments, 
from simple rooms to complex multi-story buildings, 
and convert them into knowledge graphs to answer competency questions. 
By integrating scene graphs with a robot ontology, we aim to transform abstract scene data into actionable knowledge for robots, enabling them to interpret and interact with their environment.

Addressing the challenge of scene description format incompatibility, we propose a novel pipeline that seamlessly integrates with existing workflows, allowing for environmental data in any format to be machine-readable and understandable. The methodology involves translating scene graphs from different scene description formats into Universal Scene Description (USD) format \cite{elkoura2019deep} and enriching the scene graphs with semantic annotations derived from semantic reporting. These graphs are semantically annotated and mapped into a knowledge graph, which the robots use in household manipulation tasks. Competency questions are employed to validate that the knowledge within the ontology aligns with the intended tasks \cite{winieswski2019}.

The scientific contributions of this work are manifold.
\begin{enumerate}
    \item We formalize a scene graph model capable of translating scene graphs from any common scene description format into one unified USD format (section \ref{sec:USD}).
    \item We introduce a web-based visualization tool for annotating semantic labels on USD scene graphs. This tool seamlessly integrates with a semantic reporting system, enabling semantic tagging with concept classes sourced from various ontologies (section \ref{sec:SemanticReporting}).
    \item We demonstrate that this semantic map can be utilized to answer semantic queries necessary for a table-setting task,
    showcasing the practical applications of our integrated approach (section \ref{sec:Experiment}).
\end{enumerate}

The potential impact of validating this research hypothesis is significant, as it would outline the necessary steps to transform available scene graph data structures into functional environmental models for robotic agents. Furthermore, successful validation would pave the way for future research on hierarchically labeling scene graph nodes with semantic labels referencing categories in ontologies. 

\section{Related work}

In this chapter, we explore the related work across three important areas—simulation in robotics, semantic maps, and knowledge graphs and ontologies—that are foundational to our research. Research focusing on simulation in robotics illustrates how simulation environments can serve as knowledge bases for reasoning tasks. Semantic maps are explored for their role in enhancing a robot's environmental understanding and interaction, while knowledge graphs and ontologies are essential for structuring this data into actionable knowledge.


\subsection{Simulation in Robotics}

Beginning with datasets, M. Deitke et al. introduced ProcTHOR \cite{deitke2022procthor}, a framework for procedurally generating embodied AI environments.
However, ProcTHOR is mainly accessible through Unity~\cite{nicoll2019unity} and is less suited for tasks requiring complex physics interactions. A significant application of ProcTHOR is in Habitat 3.0 by Meta AI, developed by X. Puig et al. \cite{puig2023habitat}, a simulation platform for studying collaborative tasks in domestic settings. The ProcTHOR datasets can be leveraged using our pipeline to create semantic simulation environments. 

Simulated environments have been used as testbeds for commonsense inference for robots. In Housekeep~\cite{kant2022housekeep}, a robot is tasked to learn human preferences about where many objects should be stored in a tidy house. ROOMR~\cite{Weihs_2021_CVPR} requires the robot to restore the initial configuration of a simulated environment generated via AI2-THOR~\cite{kolve2017ai2}, and this task will involve planning actions such as opening/closing or stacking. In~\cite{ding2022robot}, the authors present a system in which a PDDL planner collaborates with an LLM to repair plans that encountered an error caused by an unexpected feature of an environment. Recently, visually and physically realistic simulation models from real-world images can be generated using URDFormer \cite{chen2024urdformer}. 

\subsection{USD, Semantic Digital Twins and Semantic Maps}

USD is a framework developed by Pixar for creating, manipulating, and storing complex 3D scenes \cite{elkoura2019deep}. It is the only scene description format that can be extended to include extra environmental properties, making it capable of preserving diverse attributes from various formats and converting them into a Knowledge Graph \cite{nguyen2024translating}.

Prior work has shown that \emph{semantic digital twins}, semantically annotated digital versions of real environments, which can be created by robotic agents~\cite{Beetz2022}, provide a standardised environment encoding that robots can use to perform tasks~\cite{kuempel2021SemDT} successfully. It has also been shown that the connection of semantic digital twins to actionable knowledge graphs that link environment to object and action information can additionally be used by agents to perform task variations on varying objects~\cite{KuempelAKG} reliably.

Semantic mapping is an important part of the semantic digital twins. A semantic map is a scene graph enriched with semantic information, such as object types~\cite{cogsys12semantic_mapping}.
Research has demonstrated the utility of such semantic maps in robotic reasoning about environments. Tenorth et al. \cite{knowrobmap} introduced a semantic map that allows queries for specific components, like a heating device's handle, which are integrated into the robot's planning language. Similarly, Galindo et al.~\cite{galindo2008robot} used semantic maps to help robots reason about object locations and deduce environmental attributes, such as recognizing a kitchen by the presence of appliances like fridges and microwaves. Capobianco et al. \cite{capobianco2014knowledge} load the semantic map as a Prolog knowledge base, allowing for rule-based reasoning, e.g., for type and part relationships.
Substantial advancements have been made in the field of online creation of this type of semantic map, primarily utilizing robot sensors to learn and label the environment
\cite{suenderhauf2017}.

\subsection{Knowledge Graphs and Ontologies}

Knowledge graphs and ontologies are key techniques for knowledge representation. A knowledge graph represents entities as labeled vertices and their relationships as labeled edges, with queries (e.g., in SPARQL) retrieving connected entities. Knowledge graphs may cover specific topics, e.g. WordNet for word meanings~\cite{Fellbaum1998}, or attempt a broad coverage of encyclopedic knowledge, e.g. DBPedia~\cite{dbpedia2015}, WikiData~\cite{wikidata2014},  ConceptNet~\cite{conceptnet2017}, and the Common Sense Knowledge Graph (CSKG)~\cite{ilievski2021cskg}.

Ontologies formally define concepts and relationships, often serving as schemas for knowledge graphs. They range from foundational ontologies like BFO~\cite{BFO2015} and SUMO~\cite{sumo2011} to domain-specific ones like CORA~\cite{cora2014} and SOMA~\cite{bessler21soma} for robotics. Queries in ontologies check whether concepts are satisfiable or subsumed, using reasoners such as HermiT~\cite{hermit2010} and Konclude~\cite{konclude2014} for OWL-DL.

In simple terms, ontologies define classes of object classes, and knowledge graphs store data about these instances, enabling agents to reason and act effectively in their environments.

\section{Overview}
\label{sec:Overview}

This section provides an overview of our work, covering the methodology and the pipeline architecture. We describe our theoretical concepts in converting environmental data into usable scene graphs and then translate them into knowledge graphs. Next, we outline the pipeline architecture that enables the application of this methodology, enabling the effective execution of robotic tasks from initial human commands to their completion.

\subsection{Methodology}

\begin{figure}[h]
    \centering
    \includegraphics[width=\linewidth]{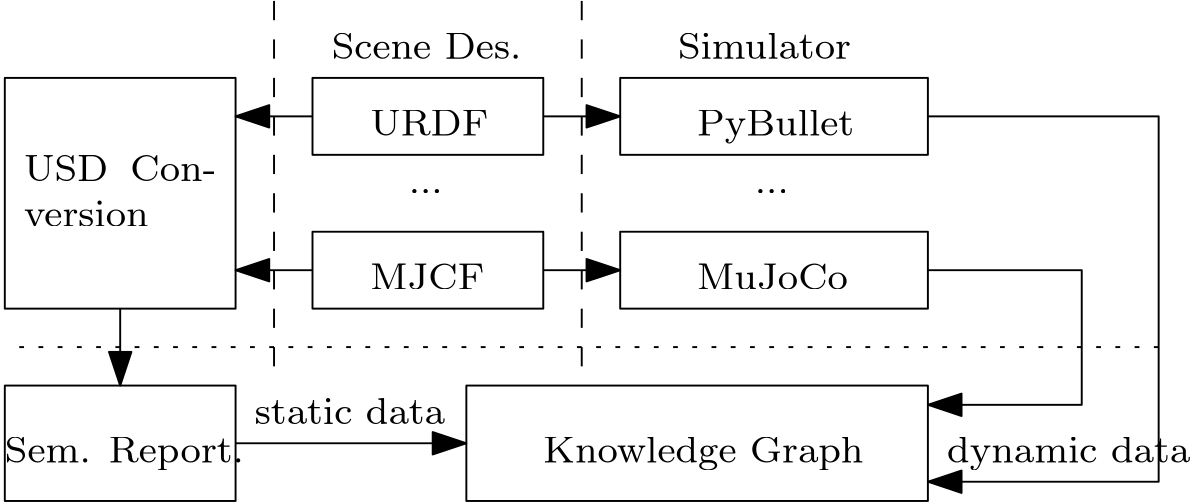}
    \caption{Diagram of the methodology}
    \label{fig:method}
\end{figure}

Our methodology starts with gathering environmental data from various sources, such as ProcTHOR, which is then combined into standardized scene graphs, which are formalized in subsection \ref{subsec:Scenegraph}. These graphs are then semantically enriched by mapping them with robot ontologies facilitated by semantic reporting. Semantic reporting automatically identifies relevant concepts from preloaded ontologies based on attributes like names or geometries in the scene graphs. Following the concepts filtered through semantic reporting, semantic labeling can be efficiently executed. Once labeled, these semantically enriched scene graphs are converted into knowledge graphs, equipping robotic agents with the necessary understanding to interact efficiently with their environment. This enriched semantic data allows robots to perform tasks more precisely and dynamically adapt to new situations. Our approach is designed for scalability and adaptability, accommodating continuous updates and refinements for complex robotic interactions (see figure \ref{fig:method}).

\subsection{Deployment of the Pipeline}

\begin{figure}[ht]
    \centering
    \includegraphics[width=\linewidth]{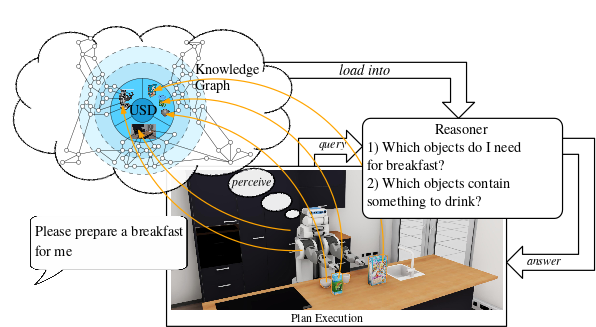}
    \caption{The pipeline's architecture starts with an ambiguous command. 
The plan execution loads environmental data, generating a USD scene graph (shown by the orange arrows), which is then transformed into a knowledge graph with multiple ontology layers. A reasoner utilizes this knowledge graph to answer competency questions, enabling real-time task execution and improving decision-making.}
    \label{fig:Architecture}
\end{figure}

The pipeline architecture described in this paper is built upon existing workflows. It starts with an ambiguous command such as "Please prepare a breakfast for me," which a natural language processor converts into executable tasks. These tasks are executed by the plan execution, which loads environmental data, including the robot and its surroundings, to create a scene graph. This scene graph is subsequently transformed into a knowledge graph and fed into a reasoner to enable essential competency questions for task execution in real time. The process is depicted in figure \ref{fig:Architecture}, and the main components involved—the plan execution, the reasoner, and the knowledge representation—are briefly introduced in the following subsections.

\subsubsection{Plan Execution}
The CRAM architecture \cite{beetz2023CRAMCognitiveArchietecture, Beetz2010CRAMPaper}, with its core component PyCRAM \cite{dech2024pycram}, provides a framework for defining generalized action plans. PyCRAM uses designators to describe objects, locations, or actions, which can be dynamically adjusted during runtime. This adds flexibility, allowing plans to adapt to different situations. While being understandable, the plan is not fully deterministic. For example, a cup's location might be unknown or changed due to human interaction. By integrating a reasoner, the cup's designator can be dynamically enriched with accurate location data and additional information, such as its contents. This real-time enrichment enables the agent to make informed decisions based on the current state of the environment.

\subsubsection{Reasoner}

The plan execution utilizes KnowRob \cite{knowrob1,knowrob2} as a knowledge representation and reasoning mechanism. KnowRob is an established knowledge representation and reasoning framework for robots. It can import ontologies, thereby enabling semantic reasoning over the embedded data. KnowRob employs a query language that is close to declarative syntax and semantics. Additionally, KnowRob allows for the easy embedding of a reasoner using a defined interface, such as in Python. In this paper, we will use KnowRob to import our custom-developed ontologies, facilitating the extraction of data required for the informed execution of robotic plans. Furthermore, we will add a reasoner specifically developed to reason about the usage of objects within DFL-based ontologies.

\subsubsection{Knowledge Representation}

The Socio-physical Model of Activities (SOMA) \cite{bessler21soma} is designed for autonomous robotic agents to understand better and execute everyday activities that involve physical and social contexts. 
SOMA is integrated into the plan execution, making knowledge from ontologies actionable by facilitating the translation of high-level semantic task descriptions into executable robot plans. 
Extensions of SOMA exist, such as SOMA\_HOME and SOMA\_DFL~\cite{SOMADFL2022}, which contain concepts for many objects that may be found in a house, such as various tools, rooms, or food items, together with information about parts, materials an object may be made of, and its typical uses. These extensions have been created through a mix of expert curation and linking with other knowledge repositories on the web, such as ConceptNet~\cite{conceptnet2017} and the Commonsense Knowledge Graph (CSKG)~\cite{ilievski2021cskg}.

The following sections will explain how to translate a scene graph into a knowledge graph. This includes parsing common scene description formats into USD and tagging semantic labels derived by semantic reporting using a tool developed in this paper.

\section{Universal Scene Description Parser}
\label{sec:USD}

Integrating diverse scene descriptions into various simulation environments and their connection to knowledge graphs necessitates a translation mechanism. We streamline this process by employing USD as an intermediary format, minimizing the coupling of different parsers and enhancing consistency. Thus, only a singular importer to an exporter from USD is required for each scene description format, as shown in figure \ref{fig:MultiverseParser}. This section outlines the formalization of robotic scene graphs focusing on the representation of rigid bodies, facilitating the mapping of standard scene description formats into USD, thereby preparing it for subsequent translation steps. Before export, the scene graph is refined to maximize its utility.

\subsection{Formalization of robotic scene graphs}
\label{subsec:Scenegraph}

The formalization begins by defining a world within a set of all possible worlds $w \in \mathcal{W}$ as a composite of bodies and global properties $w = \{B, P_{\mathrm{W}}\}$. Each body $b \in \mathcal{B}$ within this world, belonging to the set of all possible bodies $\mathcal{B}$, is described as a tuple $b = (B, J, G, P_{\mathrm{B}})$, where $B$ encapsulates child bodies attached to the body $b$, $J$ enumerates the body's joints, $G$ represents associated geometries and $P_{\mathrm{B}}$ captures the body's properties, 
such as transformation and dynamic parameters, 
which is described below.

Child bodies and joints are linked to a body, ensuring synchronized movement. For instance, like how a door handle or hinge is affixed to a door and its frame, these elements maintain their attachment with the primary body. A joint $j \in \mathcal{J}$ within $J$ is defined as $j = (t_j, b_{\mathrm{parent}}, b_{\mathrm{child}}, P_{\mathrm{J}})$, with $t_{\mathrm{J}} \in \{\mathrm{fixed}, \mathrm{revolute}, \mathrm{prismatic}, \mathrm{spherical}\}$ classifying the joint type, $b_{\mathrm{parent}}$ and $b_{\mathrm{child}}$ establishing the joint's connectivity and $P_{\mathrm{J}}$ outlining the joint's specific properties, which is described below. This structure accommodates both tree and loop kinematic configurations. 


Each geometry $g$ within the set of all possible geometries $\mathcal{G}$ is defined by the tuple $g = (t_{\mathrm{G}}, P_{\mathrm{G}})$, which comprises the geometry's type $t_{\mathrm{G}}$, selected from ${\mathrm{cube}, \mathrm{sphere}, \mathrm{cylinder}, \mathrm{mesh}}$, and its characteristic properties $P_{\mathrm{G}}$, which are classified and described below.

Properties across the various elements of the scene — whether they belong to the world, bodies, joints, or geometries — are defined by the tuple $(n, v)$, where $n$ represents the name of the property within a set of all possible names $\mathcal{N}$, and $v$ denotes the property's value from a corresponding set of allowable values $\mathcal{V_\mathrm{n}}$ depending on the property name. World properties might include aspects like gravity force or light intensity. Body properties encompass transformations relative to the parent body's frame and dynamic attributes such as mass, center of mass, and the inertia matrix. Joint properties vary based on the joint type, such as the axis of rotation and limits for \emph{prismatic} and \emph{revolute} joints. Geometry properties are divided into those dependent on geometry type — like vertices and faces for \emph{mesh} geometries or the radius for \emph{sphere} geometries — and those independent of type, such as the geometry's transformation relative to the body frame, its visibility, whether it can collide, or its default color in RGBD space. Geometry materials are organized into a series of triples and linked to geometry as a property.

\subsection{Storing robotic scene graphs into USD}
\label{subsec:ScenegraphIntoUSD}

\begin{figure}[ht]
    \centering
    \includegraphics[width=\columnwidth]{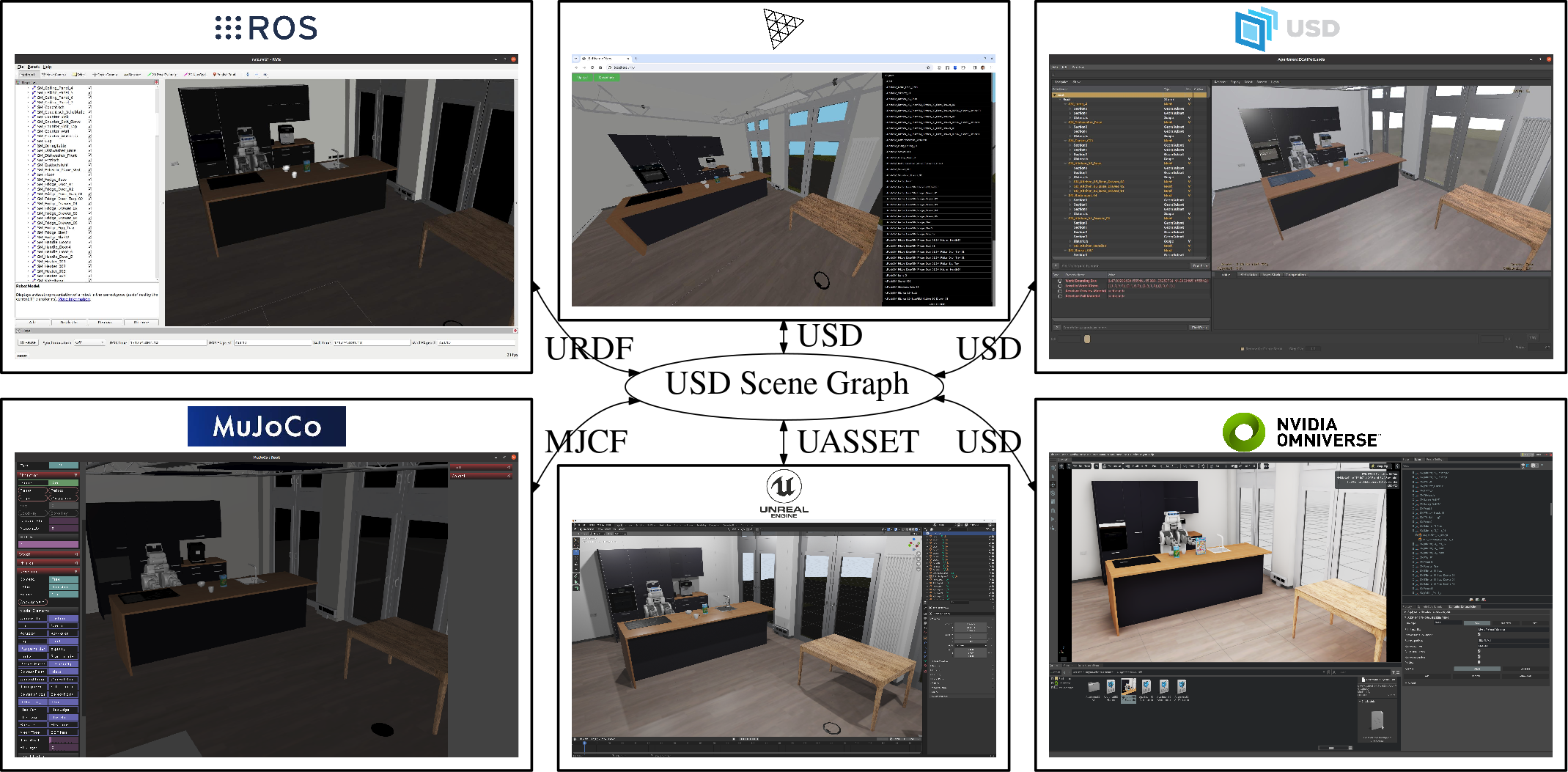}
    \caption{Storing various scene descriptions into USD and converting them into others in different simulation and visualization platforms.}
    \label{fig:MultiverseParser}
\end{figure}

Utilizing our established model, we facilitate the translation of conventional description formats like URDF, MJCF, and SDF, as outlined by M. Ivanou et al. \cite{ivanou2021robot}, into USD. Originating in XML, these descriptions are transformed into a set of triples that align with our scene graph model, which is then encoded within USD. This paper concentrates on the semantic reasoning of the scene graphs, operating under the assumption that the robot has access to all data of the environment and that each \emph{prim} (short for "primitive," which represents an individual object in USD) has a unique name, which serves as its identifier. Consequently, all lighting-related and non-rigid body data are excluded from the translation process and will be considered in future research. Our previous work \cite{nguyen2024translating} provides a detailed formalization of USD and its \emph{schemas}, facilitating a seamless and effective mapping between our scene graphs and the USD framework. Within this model, the entire world is encapsulated as the root prim, with all bodies, joints, and geometries categorized under \emph{prims} of type \emph{Xform}, \emph{Joint}, and \emph{Geom}, respectively. Their names are mapped to the USD paths, reflecting their hierarchical structures. To address the challenge of integrating unique properties inherent to each scene description format, in this paper, we introduce and integrate advanced \emph{typed schemas} and \emph{API schemas} within USD. The introduced \emph{typed schemas} serve to define new \emph{prims} that encapsulate global properties, such as the mesh and texture file paths in their original formats. These paths will be linked to the respective geometry as a property to accelerate the reverse conversion process.
Meanwhile, the \emph{API schemas} are deployed to augment the \emph{prim} with specific properties derived from the original format, which are not covered by the \emph{typed schema} of the \emph{prim}. Additionally, to accommodate more primitive geometries like \emph{ellipsoid} or \emph{capsule}, the use of \emph{sphere} or \emph{mesh} types with appropriate scaling can be effectively applied. This approach enables reciprocal translation between USD and other formats. It allows for a composite scene graph in USD, merging elements from various formats, as demonstrated in figure \ref{fig:MultiverseParser}. 

\subsection{Refinement of the robotic scene graphs}
\label{subsec:ScenegraphRefinement}


In practice, simulators should selectively utilize scene graph features, focusing on essential elements and omitting unnecessary ones. For example, RViz \cite{kam2015rviz} uses URDF for scene visualization without needing joint dynamics or physics properties, while MuJoCo \cite{todorov2012mujoco}, focused on physical interactions, excludes resource-intensive visual and non-collidable meshes. Simplifying data acquisition and computations by ignoring non-essential meshes and materials is practical when the primary focus is physical behavior rather than model accuracy. However, challenges arise when scene descriptions lack or have poorly designed dynamic properties, complicating accurate simulation outcomes.

To address these challenges, refining the scene graphs is an essential final step in the translation process. This involves improving dynamic properties through estimations based on mesh data, using B. Mirtich's method for fast and accurate polyhedral mass properties computation \cite{mirtich1996fast}. The algorithm computes mass, the center of mass, and inertia for uniform-density polyhedra by converting mass integrals into volume integrals, reducing numerical inaccuracies. Complex inertial chains are consolidated into a single root body inertia following the parallel axes theorem \cite{abdulghany2017generalization}.
When dealing with scene description formats that do not accommodate loop kinematics, problematic chains are replaced with simulator-specific alternatives, such as MJCF's 'connect' or 'weld' element. 
In the context of physical simulation and collision detection, if a mesh is overly complex, non-convex, and has a high polygon count, it needs to be decomposed into a union of convex geometries. 
Tools such as the CoACD library \cite{wei2022approximate} are available to facilitate this decomposition process. 

\section{Labeling scene graphs with semantic reports}
\label{sec:SemanticReporting}

\begin{figure}[ht]
    \centering
    \includegraphics[width=\columnwidth]{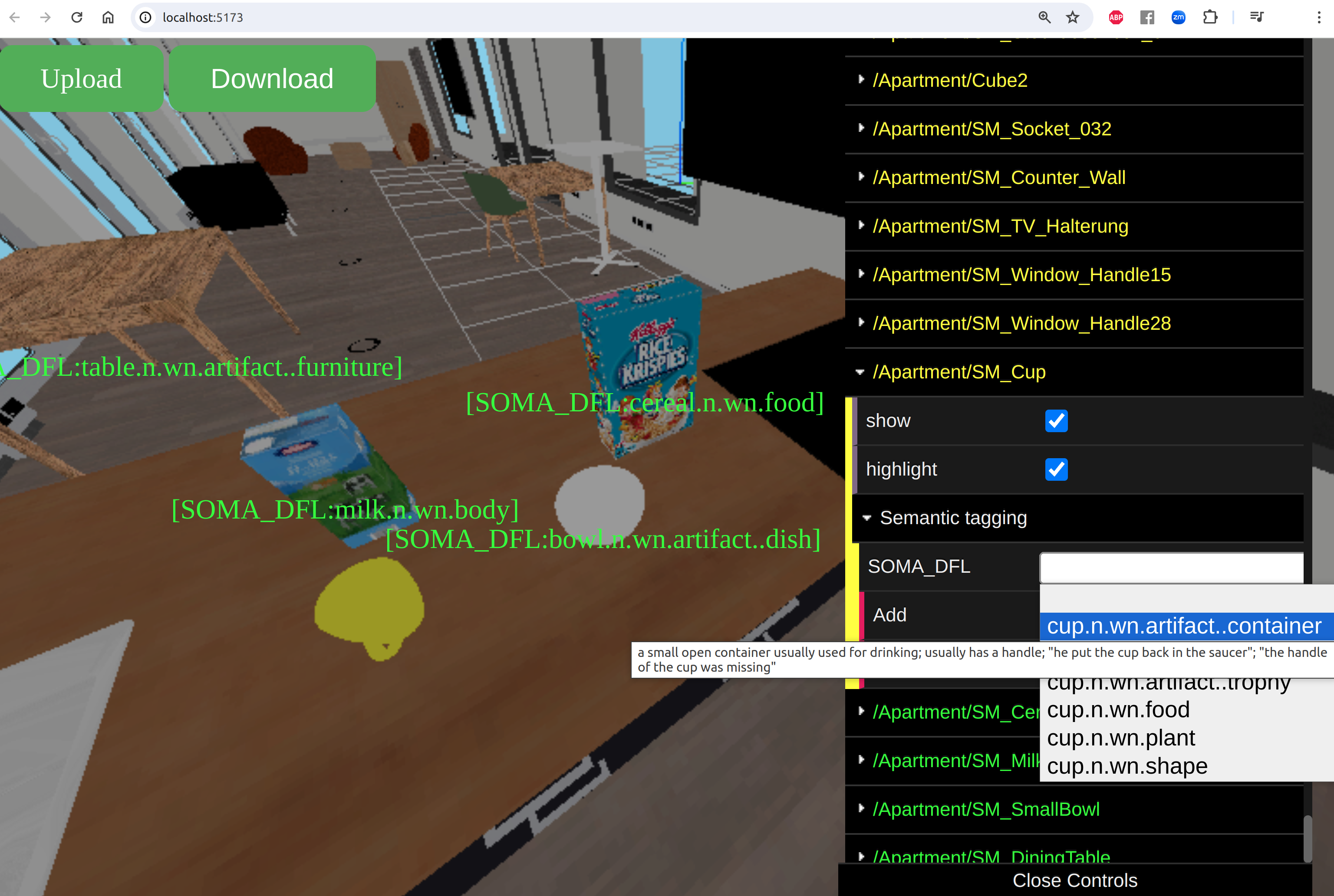}
    \caption{A snapshot of the web-based visualization tool displaying a USD scene for semantic tagging. A panel on the right side lists all the scene's \emph{prim} paths. Yellow paths indicate \emph{prims} where semantic labels from the semantic reports can be added, while green paths signify \emph{prims} that already have semantic labels added.}
    \label{fig:Multiverse-View}
\end{figure}

Building on the knowledge graph construction method outlined in \cite{nguyen2024translating}, we generate a USD layer that contains a collection of ontological concepts defined in the preloaded ontologies for semantic tagging.
Additionally, we update the schema \emph{SemanticTagAPI} with a new property, \emph{semanticTag:semanticReports}, for storing semantic reports. These enhancements facilitate semantic tagging with reduced manual effort, as explained below. The main USD scene graph, which is translated from the previous section, and the USD layer for semantic tagging are visualized using a web-based tool we developed (figure \ref{fig:Multiverse-View}). This tool employs ThreeJS \cite{danchilla2012three} for the front end and a custom-written USD parser in JavaScript, featuring upload and download capabilities. Its browser-based, backend-free architecture simplifies embedding and integration with other software, facilitating dynamic data visualization. Upon uploading a USD scene, this tool displays concept names and their definitions suggested by the semantic reporting process. It allows users to adjust semantic labels by adding or removing them before export. 

Semantic reporting attempts to automatically identify \texttt{\textbf{instanceOf}} connections between objects in a scene and classes defined in knowledge graphs such as WikiData or ConceptNet, and ontologies such as SOMA\_DFL for objects and their typical uses. This also provides further information, such as what parts or materials an object might contain or some of its typical uses and "use matches" -- situations where two objects can participate together in a task, in the roles of tool and object, acted on.

A USD prim's name and type (if given) are used to obtain the \texttt{\textbf{instanceOf}} links. We assume that a prim name -- and even more so its type -- will be strings that are informative for humans because humans will have to edit the scene, select files to import objects to the scene via the filename, etc. 

Prim names are preprocessed and sent to a text-to-triples tool, FRED~\cite{FRED}. Some context is needed to coax FRED into giving a DBPedia link; thus, if we query for some object name X, we ask FRED to parse "an X in a room" and check if what FRED identifies as the topic of this sentence has a DBPedia link. Assuming a DBPedia link is found, it is usually possible to hop from one knowledge graph to another and collect more sources of information about an object. If FRED cannot find a DBPedia link, we use a heuristic that matches the object name against a list of words associated with the concepts in SOMA\_DFL.

\begin{figure}
    \centering
    \includegraphics[width=\linewidth]{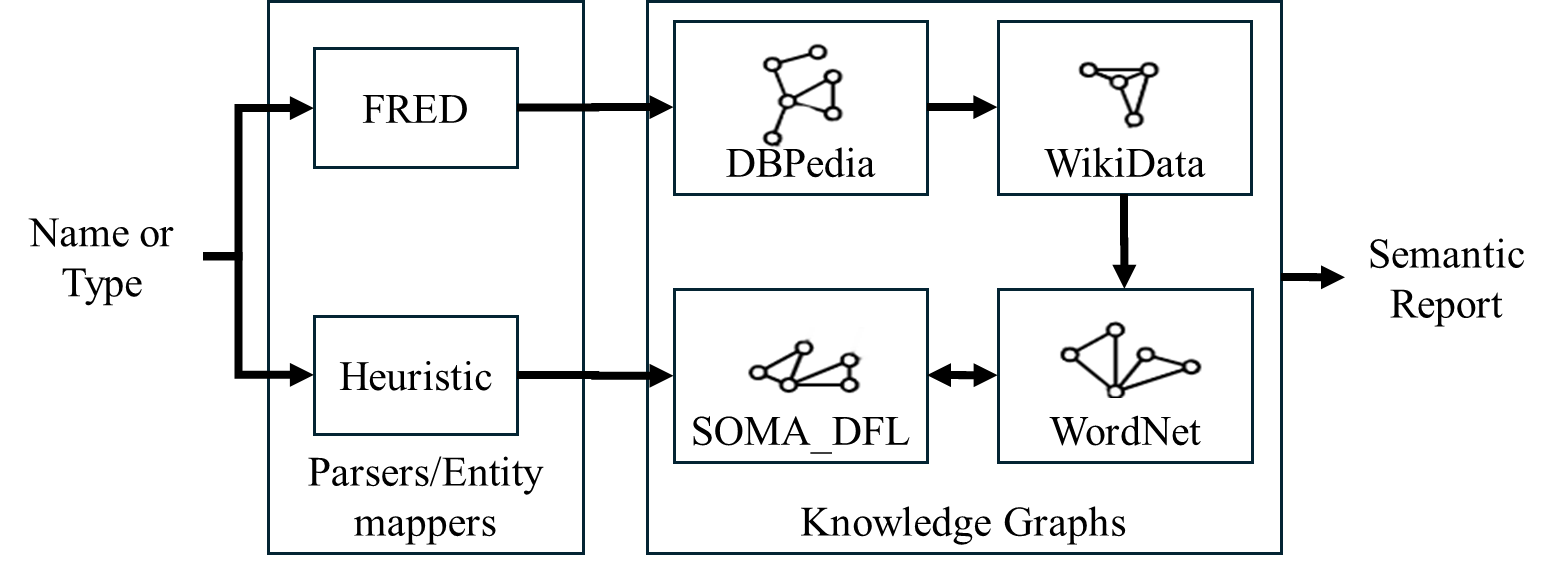}
    \caption{Overview of the semantic reporting pipeline. The text-to-triples tool FRED and our heuristics attempt to link a name or type to an entity in a knowledge graph. Arrows in the figure also show links between knowledge graphs, which allow further identification.}
    \label{fig:semrep}
\end{figure}

The above process outputs a \emph{semantic report} for a given object name or type: a list of candidates such that each candidate is a set of pairs of the form (repository, ID), where the repository is a reference to a knowledge graph or ontology, and ID is a string identifying the entity in this repository that the candidate links to. 
For each candidate, further information, such as parts, materials, and typical uses, can be retrieved. Once semantic reports for more objects in a scene are available; it is then possible to query 
for the possible use of matches between the candidates.

\section{Experiment}
\label{sec:Experiment}

To assess the validation of the hypothesis, we conducted two experiments. 
They can be replicated using a Docker container hosted on the cloud via Binderhub \footnote{\url{https://github.com/Multiverse-Framework/Multiverse-Docker/tree/IROS-2025}}. This platform-independent strategy eliminates technical setup and enhances the reproducibility of the method \cite{niedzwiecki2024cloudbased}.

\subsection{Universal Scene Description Parser}
The first experiment evaluates the effectiveness of the parser using USD as the intermediary format for translation. An apartment environment generated by ProcTHOR
is described in JSON format. The robot's description is sourced from URDF format, and object data are stored in MJCF format. The objective is to unify these scenes from various formats into a single scene in USD format, allowing them to be imported into a USD editor like Isaac Sim from NVIDIA~\cite{kainova2023overview} for further modifications. Additionally, we demonstrate the ability to convert data from USD to the original formats without losing any critical attributes, illustrating the USD's comprehensive data maintenance capabilities.

Figure \ref{fig:MultiverseParser} illustrates the actual output of the parser. The USD scene comprising the apartment, the robot, and objects from various formats was successfully created.
The ProcTHOR scene maintains its assets; any missing or inadequate dynamic properties are computed from the meshes. Currently, the USD scene from ProcTHOR remains static. Physics constraints can be added to facilitate dynamic interactions within the environment, and collision meshes can be refined for enhanced performance. Once the environment is fully developed, it can be transformed into different formats to serve various needs, with certain elements omitted to optimize performance.

In the transitivity test, we compare an open-sourced MJCF model with the one converted to USD and then revert to MJCF to verify the models' similarity. Optimizing the scene graphs by removing unnecessary elements, such as materials or textures for collision checking, while preserving key features significantly reduces resources and improves loading speeds (e.g., reducing MuJoCo scene loading from more than 10 seconds to less than 1 second without textures).

\subsection{Knowledge Graph for Symbolic Task Planning}
To demonstrate the effectiveness and robustness of our semantic map, we formulated competency questions (CQs) that apply to real robotic table-setting experiments and validate their accuracy in the given environments. The CQs are written in the KnowRob query language, enabling interactions with the ontology to retrieve relevant data. 

\begin{enumerate}[label=CQ\arabic*,leftmargin=*]
    \item Which objects do I need for breakfast?
    \item What are the storage locations for our food items?
    \item Where do we expect a tool to be?
    \item What can I grasp on an object to manipulate it?
    \item Where should I set the items for the meal?
\end{enumerate}

For illustration, queries for CQ1 and CQ4 are presented in listings \ref{lst:example}, \ref{lst:example2} in predicate logic-like pseudocode. This pseudocode shows how queries are structured in our system.

\begin{lstfloat}[ht]
		\begin{lstlisting}[tabsize=2, basicstyle=\small\ttfamily, language=python, morekeywords={subclass_of, instance_of, contains, designed_to_contain, dfl, dul, isInstanceOf, hasPart, hasPartType, isSubclassOf, useMatch, hasDisposition, instanceOf, subclassOf}, caption={CQ1: What items needed for breakfast? (stand-alone food, containers of food, or tools to eat with)}, captionpos=b, label=lst:example, frame=tb, aboveskip=0pt, framextopmargin=2pt, framexbottommargin=2pt]
?X : instanceOf(?X, 'breakfast_food'),
or (hasPartType(?X, ?C),
    subclassOf(?C, 'breakfast_food'))
or (useMatch('eat', ?X, ?Z),
    instanceOf(?Z, 'breakfast_food'))
		\end{lstlisting}
\end{lstfloat}

\begin{lstfloat}[ht]
	\begin{lstlisting}[tabsize=2, basicstyle=\small\ttfamily, language=python, morekeywords={subclass_of, instance_of, contains, designed_to_contain, dfl, dul, isInstanceOf, hasPart, hasPartType, isSubclassOf, useMatch, hasDisposition}, caption={CQ4: Where to grasp an object to manipulate it?}, captionpos=b, label=lst:example2, frame=tb, aboveskip=0pt, framextopmargin=2pt, framexbottommargin=2pt]
?X : hasDisposition(?X, 'grasp.Theme'),
or (hasPart(?X, ?P),
    hasDisposition(?P, 'grasp.Theme'))		
    \end{lstlisting}
\end{lstfloat}

Each line plays a specific role in querying our knowledge base. Lines such as \texttt{\textbf{hasDisposition}(?X, 'grasp.Theme')} look for all objects that have a particular disposition, i.e., the ability to play a specific role in a particular task, e.g., to be grasped. Lines such as \texttt{\textbf{useMatch}('eat', ?X, ?Z)} look for all objects such that \texttt{?X} is an instrument with which \texttt{?Z} is typically eaten. Some predicates link individuals to individuals, but a few, such as \texttt{\textbf{hasPartType}(?X, ?C)} link an individual \texttt{?X} to a class \texttt{?C}.
Objects can be asserted to obey \texttt{\textbf{hasPartType}} predicates based on their type, which allows the robot to query about expected parts of an object even when these are not explicitly modeled in the scene. For example, consider a scene in which there is a cereal box object, without any other object to represent the flakes inside, and the robot is asked to bring something to eat to the table. Because the robot knows cereal boxes typically contain cereal, it will retrieve this as a candidate food even when no cereal object is known.

\begin{figure}[H]
    \centering
    \includegraphics[width=0.32\textwidth]{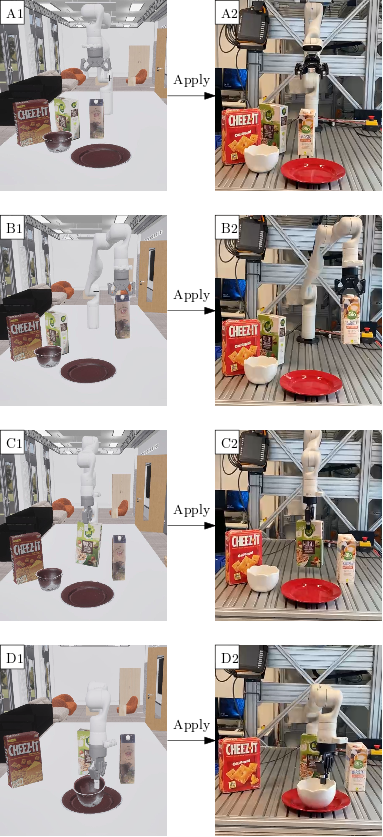}
    \caption{Knowledge Graph for Symbolic Task Planning in a Real-World Table Setting Scenario.}
    \label{fig:KinovaDemos}
\end{figure}

The competency questions can be queried in our environments, as shown in our Jupyter Notebook. To support such queries, the scene graph needs detailed semantic labels to differentiate object types (e.g., food vs. containers) and their relationships, which validates our semantic annotation.
These semantic maps provide a rich environmental representation by linking to large knowledge repositories, enabling the robot to perform tasks effectively based on the scene's configuration. 
To showcase this capability, we conducted an experiment in which the Knowledge Graph can serve as Symbolic Task planning \cite{beetz2023CRAMCognitiveArchietecture} in a real-world demo, see figure~\ref{fig:KinovaDemos}. 
After identifying and scanning all relevant objects in the scene, they are replicated in a simulation and translated into the knowledge graph. Leveraging prior domain knowledge, CQs are sent to KnowRob to answer the necessary queries for executing a household task. In this example, the task ``prepare a breakfast'' is resolved by decomposing it into three subtasks: retrieving the milk box (A1 \& B1), cereal box (C1), and bowl (D1). The robot's trajectories are generated with background knowledge using PyCRAM, simulated, and then applied to a real robot. A video of this experiment can be seen in the supplementary material.

Our rule-based reasoner consistently produces the same output for a given input. This reproducibility and consistency of this result demonstrate that using actionable knowledge derived from a scene graph is more structured and interpretable than using a Large Language Model (LLM) for symbolic reasoning and task planning. By explicitly representing relationships and dependencies between entities, knowledge graphs provide greater consistency, traceability, and verifiable reasoning. Unlike LLMs, which rely on probabilistic text patterns, knowledge graphs integrate domain-specific knowledge efficiently, minimizing hallucinations and ensuring reliability in real-world applications. However, a limitation of the actionable knowledge graph is its reliance on predefined ontological concepts as prior domain knowledge. Future work will explore the automatic generation of knowledge graphs using LLMs to address this constraint.


\section{Conclusion}
In this paper, we introduced a pipeline designed to translate environmental data from various formats into actionable knowledge for robots, enriching this data with knowledge from different sources. The pipeline operates in three stages: first, it parses the environment, robot, and object descriptions into a scene graph in USD format, integrating a semantic map from preloaded ontologies; second, the scene graph is annotated with semantic labels supported with semantic reporting and visualized in a web-based tool; and third, it converts the semantic annotated scene graph into a knowledge graph to enhance reasoning capabilities. The formalization of a scene graph, which enables translation from any robot scene description format into USD, is introduced. Our evaluation in robotic simulations highlights several key outcomes: the scene description parser effectively maintains information integrity across different formats, the semantic reporting tool simplifies the semantic labeling process, and the overall pipeline proves valuable for querying in household tasks.

\section{Acknowledgement}
This research was supported by the euROBIN project, funded by the European Union’s Horizon Europe Framework Programme under Grant Agreement No. 101070596.


\bibliographystyle{amsplain}
\bibliography{references.bib}

\end{document}